\documentclass[11pt,letterpaper]{Sengoku}
\pdfmapfile{+noto.map}
\usepackage{booktabs}
\usepackage{algorithm}
\usepackage{algpseudocode}
\usepackage[section]{placeins}
\usepackage{flafter}

\title{%
  {\fontsize{22}{26}\selectfont\textcolor{SengokuNavy}{Agents' Last Doubt}}\\[5pt]
  {\fontsize{14}{18}\selectfont
    \textcolor{SengokuTitleTransitionA}{Knowing When to Act}
    \textcolor{SengokuTitleTransitionB}{and When to Step Back}\\
    \textcolor{SengokuTitleRose}{in Tool-Using Agents}\par}%
}
\shorttitle{Agents' Last Doubt}
\runningauthor{Qisai Liu et al.}
\affiliationlayout{inline}
\affiliation{iowa}{Iowa State University}
\affiliation{nd}{University of Notre Dame}
\affiliation{rutgers}{Rutgers University}
\affiliation{liquid}{Liquid AI}
\affiliation{dtu}{\mbox{Technical University of Denmark}}
\author[iowa]{Qisai Liu\textsuperscript{*}} 
\author[nd]{Qingyue Jiao\textsuperscript{*}}
\author[rutgers]{Bangwei Guo}
\author[rutgers]{Wenxi Dai}
\author[liquid]{Changyi Yang}
\author[dtu]{Jiahe Liu}
\author[rutgers]{Dimitris N. Metaxas}
\author[nd]{Yiyu Shi}
\author[iowa]{Zhanhong Jiang}
\author[rutgers]{Minghao Guo}
\hypersetup{
  pdftitle={Agents' Last Doubt: Knowing When to Act and When to Step Back in Tool-Using Agents},
  pdfauthor={Qisai Liu, Qingyue Jiao, Bangwei Guo, Wenxi Dai, Changyi Yang, JIahe Liu, Dimitris N. Metaxas, Yiyu Shi, Zhanhong Jiang, Minghao Guo}
}

\begin{document}
\begin{abstract}
Tool-using LLM agents repeatedly face decision points under uncertainty, choosing between autonomous action, additional information gathering, and human escalation. These decisions require balancing task success against interaction cost and the risk of irreversible errors. Learning to make these trade-offs is challenging because existing interaction trajectories record only the outcomes of actions taken, leaving the consequences of alternative actions unobserved. To address this gap, we construct a counterfactual dataset containing 3,115 action outcomes from 1,036 decision points in ToolSandbox. At each decision point, we execute each legal action and evaluate its downstream outcomes based on task completion, interaction cost, and irreversible errors. Using this dataset, we train \emph{SURE-Act}, a controller that selects actions for a frozen backbone agent at inference time. Our experiments show that learning to rank alternative actions improves offline action-selection utility by 0.0309 over a regression-only baseline. We further show that adding SURE-Act enables end-to-end gains of up to 0.0575 in ToolSandbox's Overall Score with the Qwen3-4B agent. Moreover, SURE-Act outperforms adapted decision rules from SAGE-Agent and Calibrate-Then-Act by 0.1506 and 0.1438, respectively. This work establishes the first counterfactual dataset for ToolSandbox and provides actionable insights into decision-making under uncertainty for tool-using agents.


\end{abstract}

\maketitle

\let\thefootnote\relax\footnotetext{\textsuperscript{*}Core contribution.}

\section{Introduction}
\begin{figure}[!htbp]
\centering
\includegraphics[width=\textwidth]{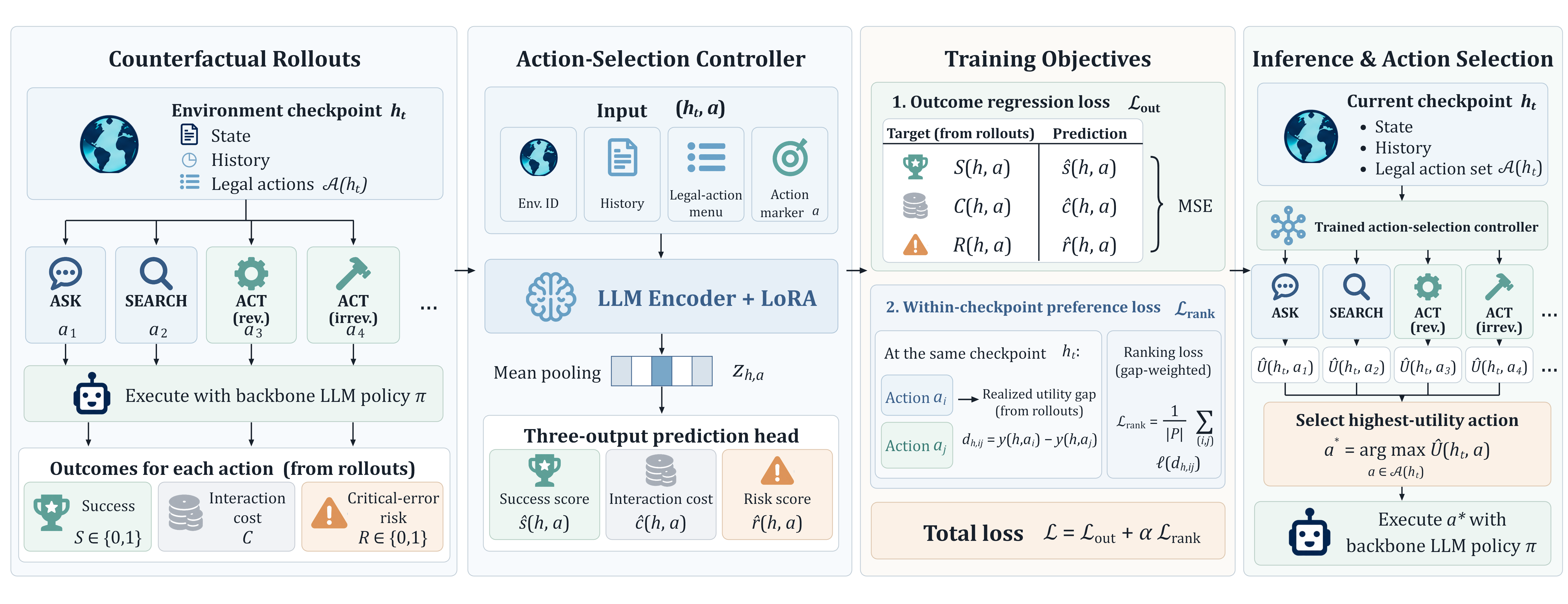}
\caption{Overview of SURE-Act. Counterfactual rollouts from an environment checkpoint evaluate
each legal action using a backbone LLM policy, producing success, interaction-cost, and
critical-error outcomes. An LLM encoder with LoRA and a three-output prediction head estimates
those outcomes from the checkpoint and the candidate action. Training combines an outcome
regression loss with a gap-weighted, within-checkpoint preference loss derived from realized
rollout utilities. At inference, the trained controller scores the legal actions, selects the one
with the highest predicted utility, and executes it using the backbone LLM policy.}
\label{fig:method}
\end{figure}

Tool-using large language model (LLM) agents complete tasks through multi-turn interactions with users and external environments \citep{yao-etal-2023-react,toolsandbox2025}. However, successful task completion alone does not establish that an agent made efficient decisions along the way. An agent may make unnecessary tool calls or ask the user for information already available in the environment. Conversely, acting before resolving uncertainty can lead to irreversible errors. At each turn, an agent must decide whether to act on the available information, gather more evidence, ask for clarification, or escalate to human assistance. Learning to make these decisions requires balancing task completion, interaction cost, and the risk of irreversible errors.

Recent work has studied the trade-off between the cost of seeking information and its value for decisions under uncertainty \citep{structuncertainty2026,cta2026,guo-etal-2026-deepsieve}. These efforts motivate a broader question: how should an agent choose among the available actions when each may lead to different outcomes over subsequent turns? Estimating uncertainty in the current state is only part of this problem. The agent must also assess whether additional information would change its decision, what obtaining that information would cost, and what could happen if it acted immediately. This motivates comparing alternative actions by their downstream consequences at the same decision point.


Obtaining supervision for this comparison remains challenging. Existing datasets contain trajectories that show the actions taken and their outcomes, but do not directly reveal what would have happened under alternative choices. A successful trajectory therefore does not establish whether another action could have achieved the same result at lower cost. Likewise, a failed trajectory does not identify which alternative action could have prevented the failure. Learning from these trajectories alone leaves the relative value of unchosen actions unexplored.

To address this gap, we introduce a framework that uses counterfactual rollouts to supervise action selection. We obtain these labels directly from the environment. At each decision point, we clone the environment state, execute each legal meta-action for one turn, and continue under the same frozen policy (Figure~\ref{fig:method}). These counterfactual branches provide an executable label for every available action at that state. A separate controller then learns to choose among those actions while the underlying policy remains unchanged. For each candidate action, the controller predicts task success, interaction cost, and critical-error risk, combines these predictions using a predefined utility function, and selects the action with the highest predicted utility. However, accurate absolute outcome predictions may still fail to capture the small utility differences that determine action rankings. To address this limitation, we directly supervise pairwise comparisons between actions at the same decision point using a ranking loss.  That loss is ordinary learning to rank applied to an unusual source. Although the loss follows standard learning-to-rank principles, its supervision comes from environment-generated counterfactual outcomes. We also establish a theoretical limitation: with only one stochastic rollout per action, observed utility differences may produce rankings that disagree with the expected utilities. We demonstrate this through a bounded counterexample (Appendix~\ref{app:theory}).



Adding the pairwise ranking loss improves action ranking over a matched regression-only objective across hundreds of family-held-out partitions. However, these partitions reuse the same family pool, and the improvement becomes less conclusive under family-level resampling. End-to-end evaluation further highlights the importance of intervention frequency, with performance closely associated with how often the controller overrides the original agent across five decision intervention-control rules.


Motivated by this observation, we introduce a simple stopping criterion: the controller remains active until it selects the same action three consecutive times, after which control returns to the original agent for the remainder of the episode. This criterion improves performance across both evaluation pools, with gains assessed against serving noise measured within each job. Among the five intervention-control rules evaluated, it is the only one that maintains a positive effect after reweighting by task family. Positive mean improvements are observed with Qwen3-4B, Qwen3-8B, and Qwen3-14B backbones (Table~\ref{tab:transfer}), and the system outperforms two adapted decision rules evaluated under the same experimental framework.

Our main contributions are the following.

\begin{itemize}
\item \textbf{Executable Counterfactual Supervision.} A procedure that makes the environment the evaluator, branching a live checkpoint across every legal action and scoring each continuation under a frozen policy. On ToolSandbox it yields 1,036 checkpoints from 378 task IDs and 82 families, carrying 3,115 scored action outcomes (Table~\ref{tab:benchmark}).
\item \textbf{Theoretical analysis of action ranking.} We establish a bound connecting within-decision-point ranking errors to decision regret and characterize how stochastic single-rollout supervision can misalign learned action rankings with expected utility.
\item \textbf{SURE-Act with Intervention-Aware Control.} We develop a controller that predicts action utility and introduce a preregistered stopping criterion that limits unnecessary interventions. The resulting approach yields positive mean improvements across the evaluated pools and three backbone sizes, and outperforms two adapted decision rules.
\end{itemize}

\section{Related Work}

\begin{table}[!htbp]
\centering
\caption{Where this work sits among methods that act under interaction uncertainty. Columns are defined in the text.}
\label{tab:taxonomy}
\scriptsize
\setlength{\tabcolsep}{2.5pt}
\resizebox{\ifdim\width>\columnwidth \columnwidth\else\width\fi}{!}{%
\begin{tabular}{@{}lcccc@{}}
\toprule
\textbf{Method} & \textbf{Frozen} & \textbf{Executed} & \textbf{Declines} & \textbf{Signal} \\
\midrule
Value-of-information clarify & \checkmark & $\times$ & \checkmark & prior \\
Budgeted env.\ queries & \checkmark & $\times$ & \checkmark & budget \\
Calibrate then act & \checkmark & $\times$ & \checkmark & threshold \\
Abstention and escalation & \checkmark & $\times$ & \checkmark & heuristic \\
Uncertainty-shaped RL & $\times$ & $\times$ & $\times$ & reward \\
Replay for blame & \checkmark & \checkmark & $\times$ & branches \\
Continuations for credit & $\times$ & \checkmark & $\times$ & branches \\
\midrule
\textbf{SURE-Act (ours)} & \checkmark & \checkmark & \checkmark & branches \\
\bottomrule
\end{tabular}%
}
\end{table}

\subsection{Agent decision-making under uncertainty}

AI agents often face uncertainty when interacting with users and invoking tools. Ambiguous user goals and missing task information can lead agents to select the wrong tools or provide incorrect arguments \cite{kirchhof2025position,structuncertainty2026}. Existing work uses uncertainty estimates to guide whether agents should invoke tool calls or ask for clarification. Building on broader interest in understanding model-internal representations \citep{jin2025exploring,han2026sage, xu2026ael, zhu2026trust}, MICE for CATs uses model-internal signals to estimate whether a tool call is correct and decide whether to execute or abstain \citep{subramani-etal-2025-mice}. Clarify When Necessary measures uncertainty across plausible user intents to determine when to ask for clarification \citep{zhang-choi-2025-clarify}, while Ask-before-Plan combines user clarification with tool-based information gathering to support planning \citep{zhang-etal-2024-ask}. For question selection, Active Task Disambiguation chooses questions that maximize the expected reduction in task ambiguity \citep{kobalczyk2025active}. SAGE-Agent models structured uncertainty over tool parameters to select informative questions while penalizing redundancy \citep{structuncertainty2026}. Existing benchmarks evaluate how agents select actions under uncertainty and use tools to complete user goals. When2Call tests whether agents should call a tool, ask for clarification, answer directly, or acknowledge that they cannot answer \citep{ross-etal-2025-when2call}. FAIL-TALMS evaluates how agents handle ambiguous user requests and unavailable tools \citep{trevino-etal-2025-benchmarking}. $\tau^2$-bench evaluates tasks that require agents and users to coordinate actions in a shared environment \citep{tau2bench2025}, while ToolSandbox evaluates multi-turn tool use with persistent environment state and dependencies between tool calls \citep{toolsandbox2025}.

\subsection{Learning for multi-turn interactions}

Improving sequential LLM behavior involves structuring reasoning processes \citep{jin2024impact,jin2025disentangling} and learning interaction strategies that improve task outcomes across successive turns. Prior work learns from preferences over complete conversations to optimize response policies \citep{shani2024multiturn}. Other studies train specific interaction strategies using feedback from simulated conversations. PPDPP trains a dialogue planner using task-oriented rewards \citep{deng2024plugandplaypolicyplannerlarge}, while STaR-GATE learns clarification questions that improve subsequent answers \citep{andukuri2024stargateteachinglanguagemodels}. Rollout-based methods evaluate individual decisions through their downstream outcomes. CollabLLM simulates future conversations to reward responses that improve task outcomes and interaction quality \citep{wu2025collabllm}. REFUEL samples alternative responses from the same conversation history and updates the policy by regressing on differences in their downstream rewards \citep{gao2025regressing}. Table~\ref{tab:taxonomy} organizes these lines of work along four axes. Frozen indicates whether the underlying policy remains unchanged, Executed distinguishes supervision from actual rollouts from supervision based on model-generated predictions, Declines indicates whether a method can choose not to act, and Signal identifies the information available at each decision point. Our work uses counterfactual rollouts to supervise a separate action selector. We execute each legal meta-action from the same decision point under a frozen continuation policy. The resulting outcomes train the selector to compare task completion, interaction cost, and critical-error risk at inference time.

\begin{algorithm}[t]
\caption{The SURE-Act pipeline.}
\label{alg:repair}
\footnotesize
\begin{algorithmic}
\Statex \textbf{Input} saved environment states $\mathcal X$, frozen data-collection backbones $\pi_x$, frozen deployment backbone $\pi$, weights $\lambda,\eta,\alpha$, threshold $k$

\Statex \textbf{Phase 1}\quad Counterfactual supervision
\For{$x\in\mathcal X$, $a\in\mathcal A(x)$}
  \State $\tau_{x,a}\gets\textsc{Roll}_{\pi_x}
  \bigl(\textsc{Force}_{\pi_x}(\textsc{Clone}(x),a)\bigr)$
  \State $(s,c,r)\gets\textsc{Grade}(\tau_{x,a})$
  \State Store context $h$, action $a$, outcomes, and missing-outcome masks
  \State $y_{x,a}\gets s-\lambda c-\eta r$ when evaluable
\EndFor

\Statex \textbf{Phase 2}\quad Controller training
\State Train $\theta$ using $\mathcal L_{\mathrm{out}}+\alpha\mathcal L_{\mathrm{rank}}$
\State Select $\theta^*$ using validation action-selection utility

\Statex \textbf{Phase 3}\quad Inference with intervention control
\State $p\gets\varnothing$, $n\gets0$
\While{episode not done}
  \State Obtain current state $x$, controller context $h$, and legal actions $\mathcal A(x)$
  \State $\hat a\gets\arg\max_{a\in\mathcal A(x)}\hat Q_{\theta^*}(h,a)$
  \State $n\gets n+1$ \textbf{if} $\hat a=p$ \textbf{else} $1$, \quad $p\gets\hat a$
  \If{$n\geq k$}
    \State Let unmodified $\pi$ complete the episode
    \State \textbf{break}
  \EndIf
  \State Execute $\hat a$ for one turn using $\pi$
  \State Advance to the next decision point
\EndWhile
\end{algorithmic}
\end{algorithm}

\section{Method}

SURE-Act trains a controller to guide a frozen tool-using LLM agent (Figure~\ref{fig:method}). First, we construct counterfactual rollouts to evaluate alternative actions from the same environment state. We then use these outcomes to supervise the controller to predict action utility and learn action rankings based on the observed context. During inference, the controller selects actions until a stopping criterion returns control to the backbone agent.

\subsection{Problem Formulation}

At decision point $t$, we define $x_t$ as the saved environment state needed to resume execution and $h_t$ as the textual interaction context observed by the controller. We define a meta-action as a high-level action category that specifies how the agent should proceed. The controller selects a meta-action from the legal set $\mathcal A(x_t)$, and the frozen backbone policy $\pi$ generates the corresponding response or tool call.

We define four meta-actions. \textsc{Ask} requests clarification or missing information from the user. \textsc{Search} retrieves information without changing the environment state. \textsc{Act-Reversible} changes the environment state in a way that can be undone. \textsc{Commit-Irreversible} makes a change that cannot be undone.

We define $Q^\pi(x_t,a)$ as the expected utility of executing meta-action $a$ for one turn at decision point $t$, then following the frozen backbone policy $\pi$:
\[
\begin{aligned}
Q^\pi(x_t,a)
&=p^\pi_{\mathrm{success}}(x_t,a)
-\lambda c^\pi(x_t,a)\\
&\quad-\eta p^\pi_{\mathrm{critical}}(x_t,a).
\end{aligned}
\]
Here, $p^\pi_{\mathrm{success}}$ is the probability of task success according to the environment evaluator. $c^\pi$ is the expected interaction cost, calculated using a fixed cost profile that assigns a cost to each type of interaction. The term $p^\pi_{\mathrm{critical}}$ is the probability that the environment accepts an irreversible action without the required authorization.

The nonnegative weights $\lambda$ and $\eta$ control the penalties for interaction cost and critical errors. Our objective is to select the legal meta-action with the highest expected utility.

\subsection{Counterfactual Data Collection}

We collect decision points from agent trajectories and save the environment state at each point. For every legal meta-action, we create a separate copy of this state and restrict the backbone’s next turn to that action category. For tool-based actions, we restrict the available tools to the selected action category and require a tool call. For \textsc{Ask}, the backbone generates a question to the user.

After this turn, we remove the restriction and let the backbone agent complete the episode. We record the resulting task success, interaction cost, and critical-error occurrence as
\[
\mathbf z_{t,a}=(s_{t,a},c_{t,a},r_{t,a}),
\]
and compute the observed utility as
\[
y_{t,a}=s_{t,a}-\lambda c_{t,a}-\eta r_{t,a}.
\]
Here, $s_{t,a}$ denotes task success according to the environment evaluator, $c_{t,a}$ is the interaction cost, and $r_{t,a}$ indicates whether a critical error occurred. Each training record contains the interaction context, candidate action, outcome vector, and observed utility. 

The final dataset contains 1,036 decision points across 378 task IDs and 82 task families, with 3,115 action outcomes (Table~\ref{tab:benchmark}). We combine records generated using Qwen3-8B and Qwen3-14B and record which backbone model generated each example. Decision points collected with different backbone models remain separate, and all action branches at a given decision point use the same frozen backbone model.

We generate a rollout per action at temperature zero. Thus, each label represents a single observed outcome, rather than the exact expected utility.
\begin{table}[!htbp]
\centering
\caption{The derived resource. Every legal action at every checkpoint was executed and scored,
over 1036 checkpoints from 378 task IDs and 82 families.}
\label{tab:benchmark}
\footnotesize
\setlength{\tabcolsep}{4pt}
\resizebox{\ifdim\width>\columnwidth \columnwidth\else\width\fi}{!}{%
\begin{tabular}{lrrrr}
\toprule
Meta-action & Records & Success & Cost & Critical \\
\midrule
ask the user & 1028 & 0.441 & 0.396 & 0.043 \\
search, read only & 1008 & 0.574 & 0.356 & 0.048 \\
act reversibly & 861 & 0.564 & 0.364 & 0.034 \\
commit irreversibly & 218 & 0.471 & 0.373 & 0.037 \\
\midrule
all & 3115 & & & \\
\bottomrule
\end{tabular}
}
\end{table}

\subsection{Utility Prediction and Ranking}

The controller takes the environment identifier, legal-action set, last twelve interaction turns, and a marker identifying the candidate action as input. We use a Qwen3-1.7B encoder \citep{qwen3} with LoRA adapters \citep{hu2022lora}, mean pooling, and a linear head to predict success, interaction cost, and critical-error scores:
\[
f_\theta(h,a)=(\hat s,\hat c,\hat r),
\qquad
\hat Q_\theta(h,a)=\hat s-\lambda\hat c-\eta\hat r.
\]
Here, $\theta$ denotes the trainable controller parameters, while the backbone model remains frozen. We use $\hat Q_\theta$ to rank actions without assuming that the success and critical-error predictions are calibrated probabilities.

Then, we train the controller to predict observed outcomes using masked squared error:
\[
\mathcal L_{\mathrm{out}}
=
\frac{1}{|\mathcal M|}
\sum_{(h,a,j)\in\mathcal M}
\bigl(\hat z_{\theta,j}(h,a)-z_j(h,a)\bigr)^2,
\]
where $j$ indexes the three outcome components and $\mathcal M$ contains the context-action-component triples with evaluable targets. Missing outcomes do not contribute to the loss.

Accurate outcome predictions alone do not ensure the correct action ranking. Therefore, we add a pairwise ranking loss to directly supervise utility differences. At a decision point with context $h$, let $\mathcal P_h$ contain legal action pairs $(a,b)$ with evaluable utilities and a positive observed gap $g_{ab}=y_a-y_b$. We define
\[
\mathcal L_{\mathrm{rank}}(h)
=
\frac{1}{|\mathcal P_h|}
\sum_{(a,b)\in\mathcal P_h}
\max\{0,g_{ab}-(\hat Q_a-\hat Q_b)\},
\]
where $\hat Q_a=\hat Q_\theta(h,a)$. This loss encourages the better observed action to receive a higher predicted score, with a margin equal to the observed utility gap. Tied actions are excluded, and decision points without a valid pair contribute no ranking loss.

The combined objective is
\[
\mathcal L=\mathcal L_{\mathrm{out}}+\alpha\mathcal L_{\mathrm{rank}},
\]
where $\alpha$ controls the contribution of ranking supervision. Setting $\alpha=0$ gives the regression-only ablation, with all other training settings unchanged. We select the trained controller based on action-selection utility on validation task families.

\subsection{Inference with Intervention Control}

At each decision point, the controller proposes the legal action with the highest predicted utility:
\[
\hat a_t\in\arg\max_{a\in\mathcal A(x_t)}\hat Q_\theta(h_t,a).
\]
The backbone executes the proposal using the same one-turn constraints as in data collection. The controller is consulted again at the next decision point.

Training labels evaluate one intervention followed by the original policy. Repeated overrides change this continuation behavior, so their cumulative effect is not directly captured by the labels. We therefore use a stopping rule based on consecutive identical meta-action proposals.

The controller maintains the previous proposal $p$ and a repetition counter $n$. A matching proposal increments $n$. A different proposal resets it to one.

When $n$ reaches the preregistered threshold $k=3$, the controller stops before executing that proposal. The unmodified backbone then completes the episode. Returning control is a stopping decision, not a fifth meta-action. Continuous intervention without this rule is evaluated as an ablation.

Algorithm~\ref{alg:repair} summarizes the procedure. $\textsc{Clone}$ copies a checkpoint, $\textsc{Force}$ executes the selected meta-action for one turn, $\textsc{Roll}$ continues under the unconstrained backbone, and $\textsc{Grade}$ evaluates the resulting trajectory.

\subsection{Theoretical Analysis}

We distinguish guarantees about expected utilities from those about observed rollout labels. Fix a checkpoint and continuation policy, and write $Q_a=Q^\pi(x,a)$. Let $a^*$ maximize $Q_a$ and $\hat a$ maximize $\hat Q_a$. If
\[
\left|(\hat Q_{a^*}-\hat Q_a)-(Q_{a^*}-Q_a)\right|
\leq\epsilon
\quad\text{for all legal }a,
\]
then $Q_{a^*}-Q_{\hat a}\leq\epsilon$. This motivates learning utility differences, but the bound is conditional on an error assumption that we do not establish for the trained controller.

For observed labels, decision regret relative to the best realized action is bounded by the sum of pairwise hinge losses. This does not imply the same guarantee for expected utility.

For a bounded random utility difference $D$, the population ranking loss can favor the sign that occurs more frequently, while $\mathbb E[D]$ also depends on outcome magnitudes. The preferred ordering can therefore disagree with the expected-utility ordering. Appendix~\ref{app:theory} provides the formal results, proofs, and repeated-rollout analysis.
\section{Experiments}

\subsection{Experimental Setup}

We evaluate SURE-Act on ToolSandbox using offline action selection and end-to-end task execution. End-to-end experiments evaluate Qwen3-4B, Qwen3-8B, and Qwen3-14B backbones, including a separate pool of 89 unseen task families.

For controller training, we split data by task family into 60/20/20 training, validation, and test sets. All decision points and task variants from a family remain in the same split. We select models using validation action-selection utility and use identical splits for paired comparisons.

Offline evaluation measures the observed utility of the selected actions. End-to-end evaluation uses ToolSandbox's Overall Score. We compare SURE-Act with the unmodified backbone, controller ablations, and adapted decision rules from SAGE-Agent \citep{structuncertainty2026} and Calibrate-Then-Act \citep{cta2026}. The latter comparisons evaluate the adapted rules within our framework, rather than the complete published systems.

We report paired performance differences and distinguish split-level standard errors from family-clustered confidence intervals. Repeated splits reuse the same family pool, so their standard errors describe sensitivity to the split rather than uncertainty over new task families. The details can be found in Appendix~\ref{app:experimental_setup}.

\subsection{End-to-End Performance}

We first evaluate SURE-Act's performance on complete task execution. On test splits drawn from the 82-family development pool, SURE-Act improves ToolSandbox’s Overall Score by 0.0344 with the frozen Qwen3-8B backbone (SE 0.0082), with positive improvements on 20 of 22 evaluation partitions. A repeat of the same experiment yields a similar gain of 0.0386 (SE 0.0093).

Table~\ref{tab:transfer} reports results on 89 unseen task families across three backbone sizes. SURE-Act achieves mean improvements of 0.0575, 0.0216, and 0.0252 with Qwen3-4B, Qwen3-8B, and Qwen3-14B, respectively. These results show positive average effects across the evaluated backbones. However, the family-clustered confidence intervals for the 4B and 8B comparisons include zero, so the gains remain uncertain across task families. These score improvements do not imply improved safety: critical-error occurrence increases in every repaired evaluation for which component-level statistics are reported. Additional statistics, including family-clustered intervals, critical-error outcomes, and no-stopping comparisons, appear in Appendix~\ref{app:end_to_end_statistics}.

\begin{table}[!htbp]
\centering
\caption{The repaired controller across four evaluation settings. Each row includes the serving-noise floor from its evaluation job.}
\label{tab:transfer}
\footnotesize
\setlength{\tabcolsep}{3pt}
\resizebox{\ifdim\width>\columnwidth \columnwidth\else\width\fi}{!}{%
\begin{tabular}{lrrr}
\toprule
Population & Gain & Better & Floor \\
\midrule
Development pool, 8B & +0.0344 {\scriptsize $\pm$ 0.0082} & 20/22 & +0.0068 \\
Unseen families, 8B & +0.0216 {\scriptsize $\pm$ 0.0093} & 18/21 & +0.0011 \\
Unseen families, 4B & +0.0575 {\scriptsize $\pm$ 0.0095} & 20/22 & -0.0012 \\
Unseen families, 14B & +0.0252 {\scriptsize $\pm$ 0.0101} & 17/20 & +0.0013 \\
\bottomrule
\end{tabular}
}
\end{table}

In a separate matched comparison, SURE-Act outperforms the adapted SAGE-Agent \citep{structuncertainty2026} and Calibrate-Then-Act \citep{cta2026} decision rules by 0.1506 and 0.1438, respectively (Table~\ref{tab:loop}). The corresponding family-clustered 95\% confidence intervals are $[0.0667,0.2433]$ and $[0.0583,0.2367]$. Both exclude zero. These comparisons evaluate the adapted decision rules under our shared execution framework.

\begin{table}[!htbp]
\centering
\caption{End to end on ToolSandbox's Overall Score, as the difference from the unmodified agent. Every rule was scored in one job and paired per scenario.}
\label{tab:loop}
\footnotesize
\setlength{\tabcolsep}{3pt}
\resizebox{\ifdim\width>\columnwidth \columnwidth\else\width\fi}{!}{%
\begin{tabular}{lrrrrr}
\toprule
Rule & score & vs agent & better & rate & ours over it \\
\midrule
\textbf{SURE-Act} & 0.451 & +0.0299 {\scriptsize $\pm$ 0.0104} & 17/22 & 0.1917 &  \\
SAGE-Agent & 0.300 & -0.1224 {\scriptsize $\pm$ 0.0142} & 1/22 & 1.0000 & \textbf{+0.1506} {\scriptsize 22/22} \\
Calibrate-Then-Act & 0.307 & -0.1152 {\scriptsize $\pm$ 0.0143} & 0/22 & 1.0000 & \textbf{+0.1438} {\scriptsize 21/22} \\
agent run twice & 0.418 & -0.0051 {\scriptsize $\pm$ 0.0033} & 11/22 & 0.0000 &  \\
\bottomrule
\end{tabular}
}
\end{table}

\subsection{Effectiveness of Controller Training}

We evaluate the ranking objective through a matched offline ablation. Across 330 family-held-out partitions, adding the ranking loss improves action-selection utility by 0.0309 over the regression-only controller (SE 0.0028), with positive differences on 250 partitions. Both controllers use the same data splits and training settings apart from the ranking-loss weight. Predicting success, cost, and critical-error outcomes separately also improves utility by 0.0082 over predicting a single utility score (SE 0.0016). These comparisons support the two components of the training objective. Detailed ablations are in Appendix~\ref{app:controller_training}, Table~\ref{tab:ablation}.

In a separate evaluation over 120 partitions, the controller exceeds the best fixed-action baseline by 0.0092 (SE 0.0016) and the strongest prompted baseline by 0.0107 (SE 0.0022). These standard errors describe sensitivity to partitions of the same family pool. The family-clustered intervals for the controller-versus-baseline comparisons include zero, and the controller does not establish an offline selection advantage on the 89 unseen families. The offline results therefore support improved selection within the evaluated development pool, rather than a general improvement on new task families. Appendix~\ref{app:controller_training}, Tables~\ref{tab:baselines_full} and~\ref{tab:headline}, reports the detailed comparisons.

\subsection{Interaction Cost}

Improving the task score is only useful if the agent does not become more expensive to interact with. We therefore measure what each decision rule costs the agent it modifies, paired per scenario within one job. SURE-Act overrides at 0.192 of decision points, adds 0.133 turns, and reduces the discounted interaction cost by 0.068. The two adapted rules override at every decision point, lengthen the episode by 0.543 and 0.492 turns, and raise the cost by 0.080 and 0.077. Running the unmodified backbone twice moves the cost by less than 0.001, which bounds the serving variability of this measurement. SURE-Act is therefore the only rule in this comparison that reduces discounted interaction cost, despite slightly increasing the episode length by 0.133 turns on average, and it does so at roughly a fifth of the override rate of the adapted rules. Table~\ref{tab:cost} reports the full comparison.

\begin{table}[!htbp]
\centering
\caption{What each rule costs the agent it modifies, paired per scenario against the
unmodified run. Only the repaired rule shortens the episode's cost rather than adding to it.}
\label{tab:cost}
\footnotesize
\setlength{\tabcolsep}{4pt}
\resizebox{\ifdim\width>\columnwidth \columnwidth\else\width\fi}{!}{%
\begin{tabular}{lrrr}
\toprule
Rule & Rate & Turns & Cost \\
\midrule
SURE-Act & 0.192 & +0.133 & -0.068 \\
SAGE-Agent & 1.000 & +0.543 & +0.080 \\
Calibrate-Then-Act & 1.000 & +0.492 & +0.077 \\
agent run twice & 0.000 & -0.004 & -0.000 \\
\bottomrule
\end{tabular}
}
\end{table}

\subsection{Sensitivity to the Stopping Threshold}

The stopping rule hands the episode back after the controller repeats one choice $k$ times, and $k$ was fixed at three before the closed-loop runs. We sweep $k$ from two to five as four variants within one job per pool to test whether the result depends on that choice. On the development pool the gain is 0.0376, 0.0358, 0.0353, and 0.0119 for $k$ of two through five, and the override rate climbs from 0.089 to 0.439. On the 89 unseen families the gain is 0.0355, 0.0225, 0.0044, and $-0.0072$ over the same range, with the rate climbing from 0.096 to 0.418. The ordering is monotone on both pools, the gain falls as the threshold admits more overrides, and the registered $k$ of three clears its own serving-variability reference on both. The window of workable thresholds is narrower on unfamiliar tasks, where only two and three clear. Table~\ref{tab:latchk} reports every threshold with its interval.

\begin{table}[!htbp]
\centering
\caption{The handover threshold swept in one job per pool. Turns is the change in episode
length against the unmodified agent, and the noise floor is that agent run twice.}
\label{tab:latchk}
\footnotesize
\setlength{\tabcolsep}{4pt}
\resizebox{\ifdim\width>\columnwidth \columnwidth\else\width\fi}{!}{%
\begin{tabular}{lrrrcr}
\toprule
$k$ & Rate & Turns & Gain & Interval & Better \\
\midrule
\multicolumn{5}{l}{\emph{development}} \\
2 & 0.089 & -0.091 & +0.0376 & [+0.0148, +0.0605] & 17/21 \\
3 (ours) & 0.192 & +0.281 & +0.0358 & [+0.0163, +0.0553] & 19/21 \\
4 & 0.308 & +0.677 & +0.0353 & [+0.0165, +0.0542] & 19/21 \\
5 & 0.439 & +1.064 & +0.0119 & [-0.0107, +0.0344] & 12/21 \\
floor & 0.000 & -0.008 & -0.0008 & [-0.0080, +0.0065] & 11/21 \\
\midrule
\multicolumn{5}{l}{\emph{89 unseen families}} \\
2 & 0.096 & -0.158 & +0.0355 & [+0.0184, +0.0526] & 18/20 \\
3 (ours) & 0.201 & +0.204 & +0.0225 & [+0.0024, +0.0426] & 17/20 \\
4 & 0.316 & +0.616 & +0.0044 & [-0.0138, +0.0226] & 15/20 \\
5 & 0.418 & +1.066 & -0.0072 & [-0.0247, +0.0103] & 11/20 \\
floor & 0.000 & -0.039 & +0.0008 & [-0.0022, +0.0038] & 12/20 \\
\midrule
\bottomrule
\end{tabular}
}
\end{table}

\subsection{Offline Selection Against Baselines}

The offline evaluation scores every selector against the same per-split best fixed action over the same 120 partitions in one scoring run, so the reference cancels in each paired difference. SURE-Act leads at 0.0092 (SE 0.0016) on 93 of 120 partitions, roughly four times the next selector, a cost-aware threshold on verbalised confidence at 0.0024 on 71 of 120. A multi-signal gate follows at 0.0023. Of the eighteen selectors scored, three clear the bar and fifteen fall below it, with the adapted SAGE-Agent rule at $-0.0550$ on 6 of 120 and the always-commit baseline last at $-0.2104$ on 0 of 120. The selectors that fail tend to collapse onto a single action, which a fixed-action baseline already represents. Table~\ref{tab:baselines_full} reports every selector.

\begin{table}[tb]
\centering
\caption{All 18 selectors paired against the same per-split best fixed action, over the same 120 partitions from one scoring run.}
\label{tab:baselines_full}
\footnotesize
\setlength{\tabcolsep}{2.5pt}
\resizebox{\ifdim\width>\columnwidth \columnwidth\else\width\fi}{!}{%
\begin{tabular}{lrr}
\toprule
selector & gain & better on \\
\midrule
\textbf{SURE-Act} & \textbf{+0.0092 $\pm$ 0.0016} & 93/120 \\
cost-aware verbalised & +0.0024 $\pm$ 0.0019 & 71/120 \\
multi-signal gate & +0.0023 $\pm$ 0.0019 & 78/120 \\
prompted, plain & -0.0016 $\pm$ 0.0021 & 57/120 \\
prompted, cost-aware & -0.0028 $\pm$ 0.0025 & 58/120 \\
cost-aware entropy & -0.0057 $\pm$ 0.0015 & 35/120 \\
self-consistency & -0.0073 $\pm$ 0.0015 & 6/120 \\
always search & -0.0073 $\pm$ 0.0015 & 0/120 \\
cost-aware self-consist. & -0.0073 $\pm$ 0.0015 & 0/120 \\
token entropy & -0.0094 $\pm$ 0.0014 & 8/120 \\
verbalised confidence & -0.0111 $\pm$ 0.0020 & 31/120 \\
prompted, CoT & -0.0250 $\pm$ 0.0026 & 23/120 \\
always act & -0.0394 $\pm$ 0.0032 & 0/120 \\
ask-or-act heuristic & -0.0397 $\pm$ 0.0027 & 8/120 \\
SAGE-Agent & -0.0550 $\pm$ 0.0030 & 6/120 \\
random legal action & -0.0629 $\pm$ 0.0022 & 0/120 \\
always ask & -0.1648 $\pm$ 0.0049 & 0/120 \\
always commit & -0.2104 $\pm$ 0.0039 & 0/120 \\
\bottomrule
\end{tabular}
}
\end{table}

\subsection{Effect of Intervention Control}

We isolate the stopping rule by comparing SURE-Act with the same controller applied throughout the episode. Without the rule, the intervention rate reaches 0.9273 and Overall Score falls by 0.0707 relative to the unmodified backbone (SE 0.0154). With intervention control, the rate falls to 0.2256 and the score improves by 0.0344. Thus, better offline action selection alone does not ensure better end-to-end performance.

\begin{figure}[!htbp]
    \centering
    \includegraphics[width=0.85\linewidth,height=0.42\textheight,keepaspectratio]{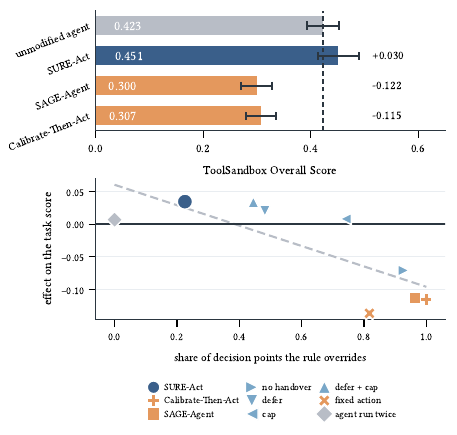}
    \caption{\textbf{Closed-loop evaluation on the development pool.} Results are averaged over 22 paired partitions. \textbf{Top:} Overall Score for each controller and the unmodified agent, with paired score differences annotated. \textbf{Bottom:} Relationship between rule effectiveness and intervention frequency across nine decision rules ($r=-0.77$). T}
    \label{fig:loop_mechanism}
\end{figure}

Figure~\ref{fig:loop_mechanism} summarizes both aspects of this result. The upper panel compares the Overall Score of each arm against the unmodified agent, while the lower panel compares intervention frequency and task performance across nine decision rules. Their correlation is $-0.77$: methods that intervene more frequently tend to perform worse in this evaluation. This association, together with the stopping-rule ablation, supports limiting repeated overrides. It does not establish intervention frequency as the sole explanation for performance.


Additional threshold and fixed-action schedule comparisons are reported in Appendix~\ref{app:intervention_control} (Table~\ref{tab:latchk}) and Appendix~\ref{app:fixed_schedule}, respectively. The latter do not resolve the learned controller's additional end-to-end benefit over a fixed-action schedule with similar intervention frequency. We therefore attribute the reported end-to-end gains to the complete SURE-Act system, which combines action selection with intervention control.

\section{Conclusion}

Interaction logs reveal only the action taken, so we generate counterfactual supervision by branching the environment, forcing each legal intervention, and scoring its continuation under a frozen policy. This supervision improves offline action selection over a matched regression objective within the evaluated family pool. End to end, however, continuous intervention lowers the benchmark score. A preregistered stopping rule returns control after three repeated choices, yielding positive partition-level mean gains across both evaluated pools. Under the same execution framework, SURE-Act also outperforms adapted implementations of the SAGE-Agent and Calibrate-Then-Act decision rules. These results suggest that, for capable tool-using agents, deciding when not to intervene matters as much as choosing which intervention to make.

\section*{Limitations}

Our evidence comes from one benchmark, one pool of 82 task families and 89 families held out from it. The offline margins hold within that pool and we report them as such. The family-clustered intervals are reported in full in Appendix~\ref{app:results}, and the scope they imply is stated where each number appears. The end-to-end finding is about a decision rule and not about the learned value model, and the Results give the two controls that bound that claim. The labels inherit the branching procedure. An intervention lasts one turn, later behaviour can undo it, each label is a single temperature-zero rollout, and the best action changed on 345 of 725 shared checkpoints across three repeated labelling runs. Finally, the environments are simulated and the repaired controller trips more critical-error conditions than the agent it improves on, so this is a task-competence result and not a safety one.

\clearpage
\appendix


\section{Experimental Setup Details}
\label{app:experimental_setup}

\subsection{Data and Evaluation Splits}

The primary ranking-loss ablation uses the counterfactual dataset described in Table~\ref{tab:benchmark}. It contains 1,036 decision points from 378 task IDs and 82 task families, with 3,115 scored action outcomes. Records are generated using Qwen3-8B and Qwen3-14B. Decision points associated with different backbone models remain separate, while all action branches at a given decision point use the same frozen backbone.

We partition task families into 60\% training, 20\% validation, and 20\% test sets. All task variants, decision points, and backbone-specific records belonging to a family remain in the same split. This prevents variants of the same family from appearing in both training and evaluation. Each split seed determines a partition by shuffling the sorted family list. Paired methods use identical seeds and therefore identical family splits.

We train a separate controller for each split and training configuration. Model selection uses validation families only. The primary ranking comparison uses 330 paired partitions of the 82-family pool. The separate offline controller-versus-baseline comparison uses 120 partitions of its development pool. These are distinct experiments, and their partition counts should not be interpreted as independent collections of task families.

End-to-end evaluation includes the development pool and a separate pool of 89 task families excluded from development. We evaluate the latter with Qwen3-4B, Qwen3-8B, and Qwen3-14B backbones. Scenario coverage and the number of completed partitions are reported for each experiment.

\subsection{Controller Training}

\begin{table}[!htbp]
\centering
\caption{Training configuration for the 120-partition controller-versus-baseline evaluation, read from the run artefacts. The 120 runs differ only in split seed. The ranking-loss ablation is a separate set of runs and uses uncentered targets.}
\label{tab:hyper}
\footnotesize
\setlength{\tabcolsep}{4pt}
\resizebox{\ifdim\width>\columnwidth \columnwidth\else\width\fi}{!}{%
\begin{tabular}{ll}
\toprule
Setting & Value \\
\midrule
Encoder & Qwen/Qwen3-1.7B \\
Epochs & 4 \\
Batch size & 8 \\
Sequence length & 1,024 \\
Learning rate & $10^{-4}$ \\
Utility weight $\lambda$ & 1 \\
Utility weight $\eta$ & 1 \\
Ranking-loss weight & 1 \\
Centred targets & yes \\
Gradient checkpointing & yes \\
Validation fraction & 0.2 \\
Test fraction & 0.2 \\
Early-stopping patience & 6 \\
Evaluate every & 50 steps \\
\midrule
\multicolumn{2}{l}{\emph{set in the training script}} \\
LoRA rank & 16 \\
LoRA alpha & 32 \\
LoRA dropout & 0.05 \\
\bottomrule
\end{tabular}
}
\end{table}

The controller uses a Qwen3-1.7B encoder with LoRA adapters, mean pooling, and a linear head that predicts success, interaction cost, and critical-error scores. Its input contains the environment identifier, legal-action set, last twelve interaction turns, and a candidate-action marker. Inputs are truncated to 1,024 tokens. Outcome labels and state hashes are excluded.

Table~\ref{tab:hyper} reports the configuration for the 120-partition controller-versus-baseline evaluation: four epochs, a batch size of eight, a learning rate of $10^{-4}$, and centered targets. The LoRA rank is 16, the LoRA scaling parameter is 32, and the dropout rate is 0.05. Gradient checkpointing is enabled. Validation is performed every 50 training steps, with an early-stopping patience of six evaluations. We select the controller checkpoint using validation action-selection utility rather than training loss.

The default utility weights are $\lambda=\eta=1$, and the ranking-loss weight is $\alpha=1$. The primary ranking-loss ablation uses uncentered outcome targets and compares $\alpha=1$ with $\alpha=0$. Both variants use the same data partitions, architecture, utility definition, and remaining training settings. The backbone models remain frozen throughout controller training.

\subsection{Baselines and Ablations}

End-to-end comparisons include the unmodified backbone, SURE-Act, and the controller without the stopping rule. The latter continues selecting meta-actions throughout the episode and isolates the effect of intervention control.

We also adapt the clarification decision rule from SAGE-Agent \citep{structuncertainty2026} and the cost-sensitive decision rule from Calibrate-Then-Act \citep{cta2026}. These rules operate within our execution framework and are evaluated on the same scenarios as SURE-Act. Each adaptation implements a selected decision rule rather than the complete system described in the original paper. The results therefore characterize these implementations under our evaluation conditions.

Offline comparisons include fixed-action rules, prompted selectors, uncertainty-based selectors, and learned controllers. The fixed-action reference is the best fixed-action baseline evaluated on each test split. Prompted selectors choose an action directly from the interaction context. Uncertainty-based selectors use signals such as verbalized confidence, token entropy, or self-consistency.

The principal training ablations remove the ranking loss or replace the three outcome predictions with a single utility prediction. Additional analyses vary the controller encoder, batch size, utility weights, and interaction-cost profile. Intervention-control analyses vary the stopping threshold and compare the learned controller with a fixed-action schedule. Each paired comparison uses the same evaluation cases and scoring convention.

\subsection{Offline Action-Selection Evaluation}

Offline evaluation measures whether the controller selects actions with higher observed utility at held-out decision points. For test set $\mathcal D_{\mathrm{test}}$, we compute
\[
U
=
\frac{1}{|\mathcal D_{\mathrm{test}}|}
\sum_{i\in\mathcal D_{\mathrm{test}}}
y_{i,\hat a_i},
\]
where $\hat a_i$ is the selected legal meta-action and $y_{i,\hat a_i}$ is its stored rollout utility.

We report gains over the best fixed-action baseline on the same test split. This reference is chosen in hindsight and is not a deployable selection strategy. However, it is shared across paired methods and cancels when computing their performance difference.

Offline evaluation uses stored counterfactual outcomes and does not require new environment execution. It measures action selection at fixed decision points, but does not capture changes to the subsequent trajectory caused by repeated controller interventions.

\subsection{End-to-End Evaluation}

End-to-end evaluation runs each method in ToolSandbox and measures its effect on complete task execution. The primary metric is ToolSandbox's Overall Score, which multiplies the milestone score by an indicator that no critical-error condition was triggered. This differs from the additive utility used to train the controller.

We compare methods on matched scenarios and report score differences relative to the unmodified backbone. We also examine milestone scores, critical-error occurrence, interaction cost, episode length, and intervention rate. Intervention rate measures how often the controller overrides the backbone's action selection.

SURE-Act uses the preregistered stopping threshold $k=3$. The controller returns control to the backbone when it proposes the same meta-action three consecutive times, before executing the third proposal. Control remains with the backbone for the rest of the episode. Threshold analyses separately evaluate $k\in\{2,3,4,5\}$.

Each end-to-end comparison includes repeated runs of the unmodified backbone within the same evaluation job. Their score difference provides a reference for serving variability. This reference helps interpret the magnitude of an observed effect, but does not replace confidence intervals or paired comparisons.

\subsection{Statistical Reporting}

For paired comparisons, we report the mean performance difference, split-level standard error, median difference, and number of partitions with a positive difference. Compared methods use the same family partitions or matched scenarios, as appropriate for the experiment.

Repeated partitions reuse a fixed pool of task families. Their standard errors therefore describe sensitivity to partition composition and should not be interpreted as uncertainty over independent samples of new families. We additionally report family-clustered estimates and 95\% confidence intervals obtained by resampling task families. These analyses preserve dependence among observations from the same family.

Results from different evaluation jobs retain their own serving-variability references and scenario coverage. Evaluation criteria and thresholds were specified before the corresponding results were inspected. Full baseline results, sensitivity analyses, and additional controls are reported in Appendix~\ref{app:results}.

\section{Formal analysis of decision-aligned ranking}
\label{app:theory}

\subsection{The training objective in full}

The base objective is masked mean-squared error over the observed outcome components, $L_{\mathrm{out}}=\sum_{h,a,k}m_{hak}(\hat y_{hak}-y_{hak})^2/\max(1,\sum_{h,a,k}m_{hak})$. Here $m_{hak}\in\{0,1\}$ indicates whether component $k$ is observed for action $a$ at checkpoint $h$, and $\hat y_{hak}$ and $y_{hak}$ are its predicted score and realized target. Let $y(h,a)$ be the realized utility from the counterfactual rollout and $\hat Q_\theta(h,a)=\hat s-\lambda\hat c-\eta\hat r$ the plug-in score from the three regression outputs. For two legal actions $a_i$, $a_j$ at the same checkpoint, define the realized gap $d_{h,ij}=y(h,a_i)-y(h,a_j)$ and predicted contrast $\hat d_{h,ij}=\hat Q_\theta(h,a_i)-\hat Q_\theta(h,a_j)$. For the set $\mathcal{P}$ of unordered legal within-checkpoint pairs in a batch, the ranking term is $L_{\mathrm{rank}} = \frac{1}{|\mathcal{P}|} \sum_{(h,i,j)\in\mathcal{P}} \max(0, |d_{h,ij}| - \operatorname{sign}(d_{h,ij})\hat d_{h,ij})$.

\subsection{Notation and assumptions}

Fix a checkpoint $h$ with $K_h$ legal actions $\mathcal A(h)$. For action $a$, let $Y^\pi_h(a)$ be the random terminal utility after forcing $a$ for one turn and then following the fixed continuation policy $\pi$. Define $Q_a=Q^\pi(h,a)=\mathbb E[Y^\pi_h(a)\mid h]$. The primary experiment observes one rollout outcome $y_a$ per checkpoint-action pair, and it does not observe $Q_a$ directly. The value $Q^\pi(h,a)$ is optimal only within the restricted class that chooses one intervention at $h$ and then follows $\pi$.

\subsection{Expected utility and decision-relevant contrasts}

\textbf{Lemma 1 (common offsets do not affect action choice).} Adding the same checkpoint-specific constant $c_h$ to every predicted action value leaves both the maximizing action and every within-checkpoint pairwise difference unchanged. For a scalar outcome head with target $z_a$ and prediction $\hat z_a$, define their action means $\bar z$ and $\bar{\hat z}$. Then

\[
\begin{gathered}
\sum_a(\hat z_a-z_a)^2
=K_h(\bar{\hat z}-\bar z)^2\\
+\sum_a\left[(\hat z_a-\bar{\hat z})-(z_a-\bar z)\right]^2.
\end{gathered}
\]

The identity separates error in the checkpoint-level outcome from error in action contrasts. The ranking term is invariant to the former. That motivates its decision alignment, but it does not show that the term is easier to optimize or that it improves generalization.

\textbf{Lemma 2 (contrast error bounds expected decision regret).} Let $a^\star\in\arg\max_a Q_a$ and $\hat a\in\arg\max_a\hat Q_a$. If

\[
\begin{gathered}
\left|(\hat Q_{a^\star}-\hat Q_a)-(Q_{a^\star}-Q_a)\right|\leq\epsilon\\
\text{for every }a\in\mathcal A(h),
\end{gathered}
\]

then $Q_{a^\star}-Q_{\hat a}\leq\epsilon$. If the best-versus-second-best expected-value gap is greater than $\epsilon$, the selected action is optimal. This is conditional on a uniform contrast-error bound, and we do not establish that condition for the trained controller.

\textbf{Proof.} Since $\hat a$ maximizes $\hat Q$, $\hat Q_{a^\star}-\hat Q_{\hat a}\leq0$. Applying the assumed contrast bound to $(a^\star,\hat a)$ gives $Q_{a^\star}-Q_{\hat a}\leq\epsilon$. If a suboptimal action were selected despite a best-action gap greater than $\epsilon$, this inequality would be violated. $\square$

\subsection{Guarantees for the realized-gap hinge}

For an unordered pair $\{a,b\}$, let $d_{ab}=y_a-y_b$ and $\hat d_{ab}=\hat Q_a-\hat Q_b$. The implemented hinge is

\[
\ell_{ab}=\max\left(0, |d_{ab}|-\operatorname{sign}(d_{ab})\hat d_{ab}\right).
\]

Let $H_h=\sum_{\{a,b\}}\ell_{ab}$, summing over all unordered legal-action pairs, and let $a_y\in\arg\max_a y_a$ and $\hat a\in\arg\max_a\hat Q_a$.

\textbf{Proposition 1 (realized-label regret).} For $K_h\geq2$,

\[
y_{a_y}-y_{\hat a}\leq H_h
=\binom{K_h}{2}\bar L_h,
\]

where $\bar L_h$ is the mean hinge over those pairs. This is a checkpoint-local guarantee against the realized labels, not against $Q^\pi$.

\textbf{Proof.} If the realized-label regret is positive, the pair $(a_y,\hat a)$ has gap $g=y_{a_y}-y_{\hat a}>0$. Because $\hat a$ maximizes the predicted score, $\hat Q_{a_y}-\hat Q_{\hat a}\leq0$, so this pair's hinge is at least $g$. All other hinge terms are nonnegative. If regret is zero, the result follows from $H_h\geq0$. $\square$

\textbf{Proposition 2 (sign-frequency orientation of the implemented hinge).} For this specific realized-gap hinge, let $D=Y(a_1)-Y(a_0)$ be a bounded realized utility difference, and define the two-action population risk for predicted difference $d$ as

\[
R(d)=\mathbb E\left[\max\{0, |D|-\operatorname{sign}(D)d\}\right].
\]

Write $p_+=\Pr(D>0)$ and $p_-=\Pr(D<0)$, with $p_++p_->0$. If $p_+>p_-$, every minimizer of $R$ is positive. If $p_->p_+$, every minimizer is negative. If $p_+=p_-$, zero is a minimizer. Thus the population hinge can orient the ranking by which outcome sign occurs more often, while $\mathbb E[D]$ also depends on gap magnitudes.

\textbf{Proof.} For $D>0$, the loss is $\max(0,D-d)$ and its derivative at zero is $-1$. For $D<0$, it is $\max(0,-D+d)$ and its derivative at zero is $+1$. For $D=0$, the loss is identically zero. Boundedness permits taking the expectation of these derivatives, giving $R'(0)=p_--p_+$. The risk is convex and has a minimizer for bounded $D$. If this derivative is negative, every minimizer lies to the right of zero. If it is positive, every minimizer lies to the left. If it is zero, convexity makes zero a minimizer. $\square$

For example, let $D=0.9$ with probability $0.11$ and $D=-0.1$ with probability $0.89$. Then $\mathbb E[D]=0.01>0$, but $p_->p_+$ and the risk is minimized at $d=-0.1$. In this case,

\[
R(d)=\begin{cases}
0.11(0.9-d), & d\leq-0.1,\\
0.188+0.78d, & -0.1\leq d\leq0.9,\\
0.89(0.1+d), & d\geq0.9.
\end{cases}
\]

This loss-specific result does not constitute a general theory of stochastic ranking. Prior work studies consistency of ranking by estimated expected utility more broadly \citep{clemencon2008ranking}.

\subsection{Baseline-relative action-value targets}

The conditional-mean identity for squared loss is standard. Ranking theory has studied when sorting estimates of expected utility is consistent \citep{clemencon2008ranking}. The following statement applies that identity to a possible baseline-relative counterfactual target, and it is not the method evaluated in this paper.

Fix a checkpoint $h$, a reference action $a_0$, and a legal action $a$. Define

\[
\begin{gathered}
D_a=Y^\pi_h(a)-Y^\pi_h(a_0),\\
\Delta Q_a=Q^\pi(h,a)-Q^\pi(h,a_0)=\mathbb E[D_a\mid h].
\end{gathered}
\]

If $D_a$ has finite conditional second moment, the population squared-loss risk $\mathbb E[(f(h,a)-D_a)^2\mid h]$ is minimized by $f(h,a)=\Delta Q_a$. If estimates use $f(h,a_0)=0$ and satisfy $\max_{a\in\mathcal A(h)}|f(h,a)-\Delta Q_a|\leq\epsilon$, then choosing $\hat a\in\arg\max_a f(h,a)$ has expected-utility regret at most $2\epsilon$ relative to $a^\star\in\arg\max_a Q^\pi(h,a)$.

\textbf{Proof.} The conditional risk decomposes as

\[
\begin{gathered}
\mathbb E[(f-D_a)^2\mid h]
=(f-\mathbb E[D_a\mid h])^2\\
+\operatorname{Var}(D_a\mid h),
\end{gathered}
\]

so its minimizer is the conditional mean. Since $f(h,\hat a)\geq f(h,a^\star)$ and both estimates have error at most $\epsilon$, $\Delta Q_{a^\star}-\Delta Q_{\hat a}\leq2\epsilon$. The reference value cancels, so this is the expected-utility regret. $\square$

This property does not guarantee finite-sample accuracy or generalization, and one rollout is an unbiased sample only when the rollout procedure samples the intended continuation distribution without selection bias. Temperature-zero service nondeterminism and stateful failures require separate characterization. Directly learning $\Delta Q$ is an untested follow-up, not a result about the current absolute-head model.

\subsection{Repeated rollouts and a high-probability bound}

Suppose that for a fixed checkpoint we collect $m$ independent observations of each pairwise utility difference. Assume each observation is unbiased for $Q_a-Q_b$ and has sub-Gaussian noise parameter $\nu_{ab}$. Let $\hat\Delta_{ab}$ be its sample mean, assume the estimated contrasts are score-consistent (there is a vector $\hat Q$ with $\hat\Delta_{ab}=\hat Q_a-\hat Q_b$), and let $\nu_{\max}=\max_{a\ne b}\nu_{ab}$. Set

\[
r_\delta=\nu_{\max}\sqrt{\frac{2\log(2K_h(K_h-1)/\delta)}{m}}.
\]

A union bound over the ordered action pairs gives

\[
\Pr\left[\max_{a\ne b}|\hat\Delta_{ab}-(Q_a-Q_b)|>r_\delta\right]\leq\delta.
\]

Combining this event with Lemma 2 bounds the selected action's expected-utility regret by $r_\delta$ with probability at least $1-\delta$. This requires repeated samples from the target distribution, unbiased contrasts, independent (or appropriately modeled) noise, and a valid noise-scale bound. Correlated branches affect the variance through $\operatorname{Var}(Y_a-Y_b)=\operatorname{Var}(Y_a)+\operatorname{Var}(Y_b)-2\operatorname{Cov}(Y_a,Y_b)$.

The current experiment does not meet these conditions. It uses one temperature-zero rollout per action, and repeated labeling changes the identity of the best action. The regret and concentration bounds therefore do not provide an expected-regret guarantee for the results reported here.

\section{Supplementary evaluation on tau-squared-bench}
\label{app:tau2}

This appendix reports a domain-specific evaluation we ran on a second environment and then found we could not interpret. We report it because the measurement is instructive, not because it bounds the method.

\subsection{What was run, and what it returned}

Separately trained controllers were fitted on each domain and evaluated against the per-split best fixed action. On retail, the evaluation used 736 checkpoints, 40 splits and 5,681 decisions, with a best-fixed headroom of +0.2096. On airline, it used 137 checkpoints, 40 splits and 1,090 decisions, with a headroom of +0.0967. In both domains, the controller chose \texttt{ABSTAIN\_HANDOFF} at every evaluated decision, with zero deviations and +0.0000 gain on every split. Across the two domains, there were 6,771 decisions and no deviation from the fitted fixed action.

\subsection{Why that result is about the grader, not the controller}

A uniform choice of one action across 6,771 decisions is a strong pattern, and the obvious reading is that the controller learned nothing. That reading is wrong, and the reason is in the labelling path rather than in the model.

The counterfactual labels for this environment are produced by forking an episode, forcing one action, and continuing for a fixed number of turns. The environment's own evaluator scores a simulation only when it terminates because the agent or the user stopped. Any other termination returns zero with the note that the simulation ended prematurely. A branch we truncate after a fixed number of turns never terminates naturally, so it is scored zero by construction. The single exception is the handoff action, which ends the episode by itself and therefore carries the agent-stop reason that the evaluator requires.

So in this label set, every action except handoff is assigned zero, and handoff is the only action that can score above it. A controller fitted on those labels will prefer handoff everywhere, which is exactly what we observe. The pattern is a property of how we truncated the branches, and it tells us nothing about whether the controller can select actions in this environment.

\subsection{What we are not doing about it}

We are not relabelling the branches as agent-stop to make the evaluator grade them. That would tell the grader the agent chose to stop when it did not, converting our own truncation into a claim about the agent's behaviour. That is a fabricated label rather than a measurement.

Two honest repairs exist. Each branch can be run to natural termination, paying the cost that the truncation was introduced to avoid. Or truncated branches can be graded directly on the database and action checks, and reported as partial credit rather than as the environment's reward. Neither is a small change and neither is implemented here.

\subsection{What this appendix therefore supports}

It supports a statement about our instrumentation on a second environment and nothing about the method. In particular, it is not evidence that the controller fails to transfer, and the main text does not use it as such. The transfer question on this environment remains open, and the measurement that would settle it is named above.

\section{Additional results and analyses}
\label{app:results}

\subsection{What the controller chooses, and how stable the labels are}

\begin{figure}[!htbp]
\centering
\includegraphics[width=0.85\linewidth,height=0.42\textheight,keepaspectratio]{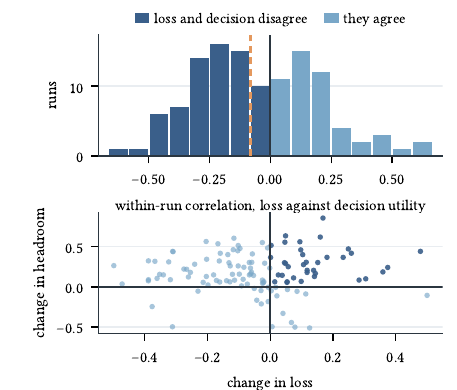}
\caption{Loss and decision quality are decoupled across 120 training runs. Above, the within-run correlation between them, median $-0.082$, with dark bars marking the runs where the two disagree. Below, how each run moved from its first evaluation to its last, with the 36 dark points improving their decision while the loss rose.}
\label{fig:training}
\end{figure}

Figure~\ref{fig:training} summarizes the within-run correlation between loss and decision utility, and the change in loss versus the change in headroom share, across 120 training runs. The action distribution bounds what the offline numbers can mean. On natural ToolSandbox checkpoints, the controller selects ASK on 0.36\% of decisions, and almost every choice falls between a read-only lookup and a reversible act. We also built a source-shift diagnostic, in which the same missing fact moves between the environment and the user's private knowledge, so the preferred action should reverse. Across 24 matched pairs (48 variants), the controller ranks probing above asking in both conditions, getting the source-dependent choice right on 0 of 24 pairs. It does not use the source of the missing information.

Labels are not stable across labelling runs. Repeating the same temperature-zero process changes the identity of the best action on 345 of 725 shared checkpoints, a flip rate of 0.476, over three repetitions differing only in the random seed. This is not an error bar on the paired ablation, whose two variants share one label draw. It does mean absolute action values do not transport between label-generation runs. Two things survive that instability. The best fixed action over the pool is identical in all three repetitions, so the reference the controller is scored against is reproducible. The aggregate margin also reproduces across those draws, as reported in the Results.

A label also measures one turn of intervention followed by the original continuation policy, which can recover later and shrink the measured difference. These are consequences of an intervention under this policy, not values of committing to an action for an episode.

\subsection{Where the offline deficit comes from}

Table~\ref{tab:baselines_full} gives the offline comparison in full, every selector against the same per-split reference. The deficit is concentrated in one confusion. 48.7\% of the per-checkpoint gap sits at checkpoints where a read-only lookup was best, and there the controller chose a reversible state change on 53 of 53 occasions. A further 44.8\% sits where asking was best, which the controller never chooses.

The state the controller reads cannot express that distinction. A term-frequency bag-of-words logistic regression, with representation and hyperparameters fixed in advance, was fitted to the same rendered text. Over 20 family-held-out partitions, it scores every action at chance, 0.4564 for read-only lookups with a family interval of [0.4093, 0.5026] and 0.5204 for asking with [0.4318, 0.6181]. The trained value head beats it on every action, 0.5344 and 0.6416 against 0.4564 and 0.5204, so the text is not information-free. The honest statement is that the distinction is not linearly recoverable from it.

Adding the missing input did not help. A preregistered feature supplying each called tool's mutating or reversible class left the registered primary unchanged and significantly degraded reversible-act discrimination, -0.0326 [-0.0496, -0.0156], better on 4 of 20 splits.

\subsection{Why the margin does not reach new families}

\begin{figure}[!htbp]
\centering
\includegraphics[width=0.85\linewidth,height=0.42\textheight,keepaspectratio]{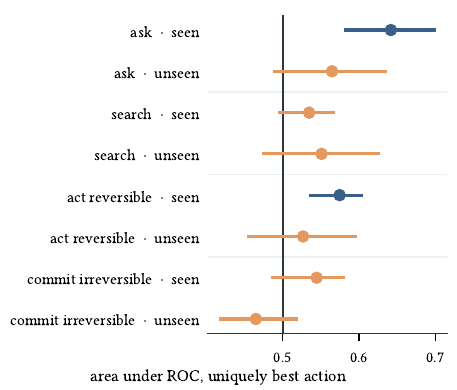}
\caption{\textbf{Within-checkpoint action discrimination.}
Area under the ROC curve (AUROC) for identifying the uniquely best action at each checkpoint. Seen and unseen correspond to 82 training families and 89 held-out families. Bars show family-clustered confidence intervals, with dark indicating intervals entirely above chance and amber indicating intervals containing chance. Three of four actions exceed chance on seen families, while none do so on unseen families.}
\label{fig:discrimination}
\end{figure}

That offline margin is stable under relabelling. Rescoring the same controllers under three independent label draws gives +0.0092, +0.0101, and +0.0131, all with positive medians.

It does not reach new task families, and we make no claim that it does. The 120 partitions reshuffle one pool of 82 families, so a standard error across them describes sensitivity to the split rather than a population, and every preregistered family-bootstrap interval includes zero. Run unchanged on 89 families never seen in any form, the controllers give -0.0000 (SE 0.0003). None of their four action rankings beat chance there. Appendix~\ref{app:results} gives the direct test, the discrimination behind it, and where the deficit sits.

The margin is at least stable under relabelling. Rescoring the same controllers under three independent label draws gives +0.0092, +0.0101, and +0.0131, all with positive medians. The offline gain reproduces across draws even though the per-checkpoint supervision beneath it does not.

Generalization to new task families is a different question, and the evidence does not support a claim. The 120 partitions are reshuffles of a single pool of 82 task families, 16 reported per partition. A standard error across partitions therefore describes sensitivity to which families are reported, not a population of families. Every one of the three preregistered family-bootstrap intervals includes zero and is approximately six to eight times as wide as its normal-approximation partition-level interval (Appendix~\ref{app:results}). These margins are therefore not established for new task families, and the pool was used throughout development, so even this interval is optimistic.

A direct test settles it. The same unchanged controllers, run on 89 ToolSandbox families never seen in any form (1060 labelled checkpoints, intersection verified empty), give -0.0000 (SE 0.0003), positive on 74 of 120 partitions. The obvious reply, that the new pool simply offers less, does not hold. It has less to win, with an oracle headroom of +0.0377 (SE 0.0013) against +0.1107 on the development pool. But capturing the same 8.3\% of it would predict +0.0031, which the measurement excludes by ten standard errors.

The reason is that there is nothing left to select with (Figure~\ref{fig:discrimination}). On the development families, the area under the ROC curve is informative for two of four actions, 0.6416 for asking and 0.5744 for reversible acts. On the 89 unseen families, all four intervals contain 0.5. No decision rule can turn discrimination at chance into a margin.

The deficit concentrates where a read-only lookup was best, and the cause is the state the controller reads. On 605 of 605 checkpoints where both actions are measured, the two renderings are byte-identical apart from one token, so an additive scorer cannot express the preference rather than lacking an input. Supplying the missing input did not help. A preregistered feature giving each called tool's mutating or reversible class left the registered primary unchanged and degraded reversible-act discrimination, -0.0326 [-0.0496, -0.0156]. Appendix~\ref{app:results} gives the full diagnosis.

\subsection{How much of the offline ceiling the head reaches}

\begin{figure}[!htbp]
\centering
\includegraphics[width=0.85\linewidth,height=0.42\textheight,keepaspectratio]{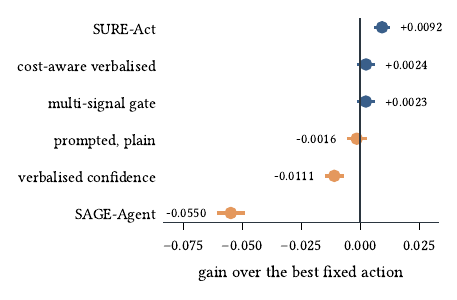}
\caption{\textbf{Offline gain over the per-split best fixed action.}
Results for six of the eighteen selectors: the three that
exceed the baseline and the three weakest of those that do not.
SURE-Act achieves the largest gain, 3.8 times that of the
next selector. Dots represent paired means, and whiskers
indicate two standard errors.
Table~\ref{tab:baselines_full} reports results for all
eighteen selectors.}
\label{fig:ceiling}
\end{figure}

Figure~\ref{fig:ceiling} ranks every selector against that bar. A perfect-utility selector exceeds it by +0.1107 (SE 0.0024), so the head captures 8.3\% of what is achievable on that pool and the published cost-aware rule 2.2\%. Against the strongest prompted baseline, the paired margin is +0.0107 (SE 0.0022) and clears the two-standard-error margin of +0.005 fixed before the run. Against the strongest verbalized-confidence baseline, the paired margin is +0.0068 (SE 0.0023), but its two-standard-error lower bound is +0.0022. It therefore does not establish an advantage exceeding the preregistered +0.005 margin under that criterion.

\subsection{Avoiding the worst action}

The head was also scored on how often it picks the single worst legal action, on checkpoints with at least three legal actions and a unique worst. 
\begin{center}\footnotesize
\setlength{\tabcolsep}{4pt}
\begin{tabular}{lrrr}
\toprule
Pool & random & the head & best fixed \\
\midrule
Development, 53 families & 0.3126 & 0.1351 & 0.1032 \\
Unseen, 80 families & 0.3085 & 0.0865 & 0.0483 \\
\bottomrule
\end{tabular}
\end{center}

Lower is better in every column. The head picks the worst action far less often than chance, by 0.1775 on the development pool and 0.2220 on the unseen families, with family-clustered intervals of [-0.2203, -0.1292] and [-0.2509, -0.1834]. It does not beat a fixed action chosen in hindsight, which is SEARCH, the worst legal action on 10.3\% and 4.8\% of those checkpoints. A rule that never moves cannot stumble onto a worse option. The registered bar was to fall below both baselines, and the registered outcome for missing it is that the account here stands unchanged, which it does. In the separate offline utility comparison, the controller gains +0.0092 over the per-split hindsight best fixed action (Table~\ref{tab:headline}). This is a different reference from the always-SEARCH rule evaluated here.

\subsection{The paired ablations in full}
\label{app:controller_training}

\begin{figure}[!htbp]
\centering
\includegraphics[width=0.85\linewidth,height=0.42\textheight,keepaspectratio]{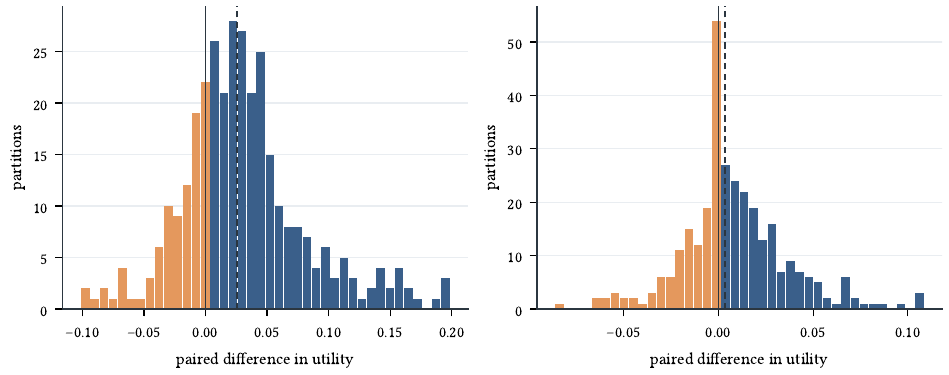}
\caption{\textbf{Training-objective ablations.}
Each bar represents the paired difference for one partition.
\textbf{Left:} Within-decision ranking versus regression-only
objective across 330 partitions (median $+0.0257$; one unchanged).
\textbf{Right:} Three outcome heads versus a single scalar
across 312 partitions (median $+0.0033$; 12 unchanged).
Dark and amber bars indicate gains and losses from the full
objective, while dashed lines mark the medians.}
\label{fig:ablations}
\end{figure}

\begin{table}[!htbp]
\centering
\caption{Paired ablations of the controller. Each row is a separate comparison on its own partitions.}
\label{tab:ablation}
\footnotesize
\setlength{\tabcolsep}{4pt}
\resizebox{\ifdim\width>\columnwidth \columnwidth\else\width\fi}{!}{%
\begin{tabular}{lrr}
\toprule
Change & Paired difference & $n$ \\
\midrule
\multicolumn{3}{l}{\emph{Design choices}} \\
\quad within-checkpoint ranking term & \textbf{+0.0309} {\scriptsize $\pm$ 0.0028} & 330 \\
\quad three outcome heads over a scalar & \textbf{+0.0082} {\scriptsize $\pm$ 0.0016} & 312 \\
\midrule
\multicolumn{3}{l}{\emph{Insensitive to}} \\
\quad a larger encoder & +0.0067 {\scriptsize $\pm$ 0.0035} & 85 \\
\quad batch size & -0.0028 {\scriptsize $\pm$ 0.0034} & 90 \\
\quad halving the cost weight & +0.0025 {\scriptsize $\pm$ 0.0050} & 41 \\
\quad doubling the cost weight & -0.0009 {\scriptsize $\pm$ 0.0054} & 44 \\
\quad halving the risk weight & +0.0039 {\scriptsize $\pm$ 0.0051} & 44 \\
\quad doubling the risk weight & +0.0038 {\scriptsize $\pm$ 0.0062} & 44 \\
\quad a cheaper cost profile & +0.0034 {\scriptsize $\pm$ 0.0049} & 47 \\
\midrule
\multicolumn{3}{l}{\emph{Sensitive to}} \\
\quad a more expensive cost profile & \textbf{-0.0205} {\scriptsize $\pm$ 0.0060} & 47 \\
\quad a different encoder family & \textbf{-0.0106} {\scriptsize $\pm$ 0.0048} & 36 \\
\bottomrule
\end{tabular}
}
\end{table}

\begin{table}[!htbp]
\centering
\caption{Controller against the per-split best fixed action over 120 family-held-out partitions.}
\label{tab:headline}
\scriptsize
\setlength{\tabcolsep}{2.5pt}
\resizebox{\ifdim\width>\columnwidth \columnwidth\else\width\fi}{!}{%
\begin{tabular}{lrrr}
\toprule
& gain & better on & $p$ \\
\midrule
\textbf{over best fixed action} & \textbf{+0.0092 $\pm$ 0.0016} & 93/120 & $<$0.0001 \\
\quad SURE-Act vs cost-aware verbalised & +0.0068 $\pm$ 0.0023 & 80/120 & 0.0003 \\
\quad SURE-Act vs best prompted & +0.0107 $\pm$ 0.0022 & 86/120 & $<$0.0001 \\
\bottomrule
\end{tabular}
}
\end{table}

Table~\ref{tab:ablation} reports eleven paired comparisons of the controller, each a separate pre-registered contrast scored on the same partitions.  Figure~\ref{fig:ablations} presents the distributions of paired differences for both objective ablations. The ranking term affects nearly every partition, whereas the three-head decomposition produces more unchanged outcomes.Table~\ref{tab:baselines_full} scores all 18 selectors against the same bar. Table~\ref{tab:headline} gives the paired controller-versus-baseline contrasts.

The ranking-loss comparison is positive on 250 of 330 partitions, with a median improvement of 0.0257. The three-outcome-versus-scalar comparison uses 312 partitions. Additional comparisons vary encoder size and family, batch size, utility weights, and interaction-cost profiles. The more expensive cost profile reduces utility by 0.0205, and changing the encoder family reduces it by 0.0106. A value head exceeds a classifier trained on the same labels by 0.0052 (SE 0.0018). In the separate 120-partition controller-versus-baseline evaluation, the gain over the per-split best fixed action has a median of 0.0086 (mean 0.0092, SE 0.0016). The gain is positive on 93 of 120 partitions.

\subsection{How much there was to win}

A perfect-utility selector exceeds the same per-split best fixed action by +0.1107 (SE 0.0024) on the same 120 partitions. The controller therefore captures 8.3\% of what is achievable and the published cost-aware rule 2.2\%. The bar is recomputed from the labels and must reproduce the scored variants' bar to within $10^{-9}$ per partition before any ceiling is reported. The headroom is not an artefact of choices that rarely matter. The action changes nothing on 12.6\% of checkpoints, and a perfect selector differs from the bar on 47.2\% of them, so roughly half the decisions carry the whole 0.1107.

Making the rare actions reachable costs utility rather than recovering it. A per-action offset or a one-versus-rest threshold, fitted on held-out validation families, raises ASK from 0.2\% to 10.8\% and irreversible commits from 0.7\% to 21.1\%. Those cost 0.0014 and 0.0066 of utility.

The two objectives also differ in shape, since one critical-error condition zeroes the native score where our utility applies a linear penalty. That difference is small. Selecting with perfect knowledge of our utility, then scoring under the native composition, beats the native best fixed action by 0.0731 (SE 0.0022) on 120 of 120 partitions. That reaches 91.5\% of the native ceiling. The shape therefore costs at most 8.5\% of what is available, and is not the closed-loop deficit.

\subsection{Additional end-to-end statistics}
\label{app:end_to_end_statistics}

This section supplements the partition-level means, standard errors, and positive-partition counts in Table~\ref{tab:transfer}. All score changes compare a method with the unmodified backbone in the same evaluation job. The primary development-pool Qwen3-8B gain is 0.0344 (SE 0.0082), with a median of 0.0342 and positive differences on 20 of 22 partitions. The independent repeat gives 0.0386 (SE 0.0093) and is a separate evaluation.

On the 89 unseen families, Qwen3-8B has a mean gain of 0.0216 (SE 0.0093), a median of 0.0392, and positive differences on 18 of 21 partitions. Qwen3-4B has a mean gain of 0.0575 (SE 0.0095), a median of 0.0720, and positive differences on 20 of 22 partitions. Their family-clustered 95\% intervals are $[-0.0211,0.0661]$ and $[-0.0063,0.1303]$, respectively. Both include zero, so positive partition-level averages do not establish gains across new task families. With Qwen3-14B, the mean gain is 0.0252 (SE 0.0101), positive on 17 of 20 partitions. No median or family-clustered interval is available in the reported statistics summarized here.

The controller without the stopping rule changes Overall Score by $-0.0707$ on the development 8B evaluation and by $-0.1909$ and $-0.1616$ on the unseen-family 8B and 4B evaluations, respectively. These are differences from the unmodified backbone, not differences from SURE-Act. They extend the stopping-rule ablation to the unseen-family evaluations. No corresponding no-stopping result is available here for 14B.

The medians and standard errors summarize variation across partitions, whereas the clustered intervals come from the task-family analysis. In particular, the development 8B family-weighted mean is 0.0296, with interval $[-0.0189,0.0804]$, compared with the partition-weighted mean of 0.0344. The clustered intervals should therefore not be interpreted as intervals computed from the partition-level standard errors. Backbone evaluations are separate jobs, and these results do not establish a monotonic relationship between backbone size and improvement.

\subsection{Stopping-threshold comparisons}
\label{app:intervention_control}

The stopping rule returns control permanently to the unmodified backbone after the controller proposes the same action three times consecutively. Table~\ref{tab:latchk} varies this threshold from $k=2$ to $k=5$ in one matched job per pool. On both the development pool and the 89 unseen families, intervention rates increase and mean score gains decrease as $k$ grows. The default $k=3$ was fixed before these comparisons. These jobs are separate from the primary end-to-end evaluation, so their gains and intervention rates should be read within each job.

Across the five intervention-control rules, the correlation between intervention rate and Overall Score is $-0.8693$. Across all nine decision rules, it is $-0.7749$. These descriptive associations do not isolate the causal contribution of intervention frequency. In the primary development evaluation, the unmodified backbone run twice differs by 0.0068 (SE 0.0037), providing a serving-variability reference for the SURE-Act gain of 0.0344. An independent repeat yields a gain of 0.0386 (SE 0.0093).

\subsection{Fixed-action schedule and outcome trade-offs}
\label{app:fixed_schedule}

A fixed-action schedule intervenes at a rate of 0.2308, close to the learned controller's rate of 0.2256, but uses the split's fitted fixed action. In the schedule-comparison job, the learned-controller rule improves Overall Score by 0.0279 (interval $[0.0076,0.0481]$), while the fixed-action schedule improves it by 0.0082 with an interval containing zero. A difference in whether two individual intervals include zero does not establish a difference between the methods. The direct paired comparisons below address that question.

The head's increment over the bare schedule is not resolved. It reads +0.0197 ([-0.0043, +0.0437], 13 of 22) and, in a preregistered replication, +0.0136 ([-0.0143, +0.0415], 10 of 20). Both point positive and neither clears zero. The quantity is a difference between two variants that are identical except at roughly one decision point in five. Its standard error is about the size of the effect, so more partitions would not settle it. The Results therefore claim the conjunction and no share for either part.

The mechanism is not improved safety. Every repaired job trips more critical-error conditions than the agent it improves on (+0.0220 development, +0.0037 unseen, +0.0115 on 4B) while winning on milestone (+0.0564 and +0.0253), and every losing rule trips fewer. The rule preserves task competence rather than safety.

We ran the same four variants on 89 ToolSandbox task families none of them has seen, verified disjoint from the development families, all 222 scenarios per partition.

One registered prediction is refuted. The reading that the mechanism is about intervention rate rather than the policy fires on the 4B job, so the account survives the policy change. But the pre-registration predicted a smaller gain on a less capable policy and the gain is larger, which it states would put the stated mechanism in trouble. What survives is the rate account rather than the preserve-a-capable-agent story.

\subsection{Scope and diagnostics of the adapted-rule comparison}
\label{app:adapted_rules}

The direct paired SURE-Act margin over the adapted SAGE-Agent rule is 0.1506 (partition-level SE 0.0186), positive on 22 of 22 partitions, with a family-clustered 95\% interval of $[0.0667,0.2433]$. Against the adapted Calibrate-Then-Act rule, the margin is 0.1438 (SE 0.0202), positive on 21 of 22 partitions, with interval $[0.0583,0.2367]$. These are the paired margins in the final column of Table~\ref{tab:loop}. Their standard errors describe the direct paired contrasts. Both clustered intervals exclude zero. This matched-rule job is separate from the transfer evaluations in Table~\ref{tab:transfer} and Appendix~\ref{app:end_to_end_statistics}.

\begin{figure}[!htbp]
\centering
\includegraphics[width=0.85\linewidth,height=0.42\textheight,keepaspectratio]{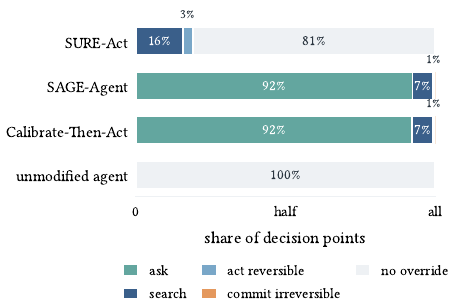}
\caption{\textbf{Decision-rule behavior across decision points.}
Distribution of five action outcomes for each decision rule.
Both ported rules override at nearly every decision point, and most overrides involve asking questions. In contrast, SURE-Act overrides at approximately one-sixth
of decision points, primarily through lookups. Outcomes accounting for less than 1\%
are labeled above the corresponding bars.}
\label{fig:behaviour}
\end{figure}

Figure~\ref{fig:behaviour} summarizes action distributions, and Table~\ref{tab:cost} reports episode-length and cost changes. SURE-Act intervenes at a rate of 0.1917 in this matched comparison. Both adapted rules intervene at every decision point. The action-distribution summary assigns the remaining 0.8083 of decision points to no override.


Each ported rule is our implementation of one portable decision rule, not the complete system the rule comes from. Those systems' published figures are reproduction targets under our scaffold rather than entries in any table here. The two rules consume the same two signals and differ only in the rule applied to them, so they are not two independent published rules agreeing.

\subsection{Weighting, intervals, and the first-decision control}

\begin{figure}[!htbp]
\centering
\includegraphics[width=0.85\linewidth,height=0.42\textheight,keepaspectratio]{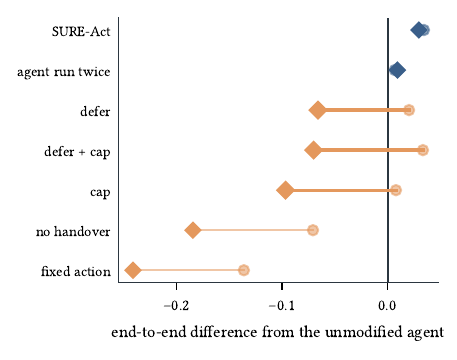}
\caption{\textbf{Partition-weighted vs.\ family-weighted effects.}
Faint circles and solid diamonds represent partition-weighted
and family-weighted effects, with connecting lines
indicating the magnitude of the weighting-induced change.
Dark indicates rules that remain positive under family weighting,
while amber indicates those that do not. Deferral, the repetition
cap, and their combination reverse from positive to negative
under family weighting (thicker lines), whereas SURE-Act is
the only repair rule that retains a positive effect.}
\label{fig:loop_weightings}
\end{figure}

Clustering on task family changes which rules survive, and only the handover rule survives. Recomputing each rule's effect one family at a time and resampling families, it reads +0.0296 against its partition-weighted +0.0344. The deferral rule, the repetition cap and their combination all flip sign (Appendix~\ref{app:results}). Its own family-clustered interval, [-0.0189, +0.0804], contains zero, so the end-to-end result carries the same scope limitation as the offline one. The mechanism claim is stronger under clustering, since intervening at every decision point costs -0.1845 family-weighted.

Recomputing each repair rule one family at a time and resampling the 82 development families, the handover rule reads +0.0296 family-weighted against +0.0344 by partition, the same sign and most of the size. The deferral rule reads -0.0660, the repetition cap -0.0967 and their combination -0.0701, the last two with family intervals excluding zero from below. The unrepaired controller reads -0.1845 family-weighted against -0.0707 by partition, and the best fixed action -0.2413 against -0.1362. Figure~\ref{fig:loop_weightings} plots the two weightings against each other for every repair rule.

A family bootstrap, with estimator and replicate count fixed in advance, gives three 95\% intervals. They are [-0.0005, +0.0374] over the best fixed action, [-0.0210, +0.0436] against the best prompted baseline, and [-0.0393, +0.0262] against the verbalized-confidence baseline. Every interval includes zero and their widths are approximately 6.0, 7.5, and 7.3 times those of the corresponding normal-approximation partition-level intervals, computed as $2\times1.96\times\mathrm{SE}$ using Table~\ref{tab:headline}. The partition-level threshold decisions were fixed against the partition-level interval and stand.

The first-decision control forces each of four fixed actions at the first decision point only, followed by the unmodified policy, and includes four unmodified-policy runs. The per-scenario best forced first action exceeds the best unmodified-policy run by 0.1500 (per-split SE 0.0084) over 21 of 21 complete partitions of the 80-family development pool. The median difference is 0.1516. Its family-clustered interval is $[0.0648,0.2286]$, and the same-policy maximum is 0.0602 (SE 0.0036). On the 89 unseen families, the four-action maximum is 0.1242 (SE 0.0019) on 22 of 22 partitions, against a same-policy maximum of 0.0363 (SE 0.0017). These hindsight maxima quantify available headroom rather than a deployable selector.

\subsection{What the family-clustered intervals imply}

The family-clustered intervals for the offline controller-versus-baseline comparisons include zero, as do the end-to-end SURE-Act-versus-backbone intervals reported for the development 8B, unseen 8B, and unseen 4B evaluations. This qualification does not apply to every contrast: the matched comparisons against the adapted SAGE-Agent and Calibrate-Then-Act rules exclude zero. Repeated partitions reuse the same family pools, so partition-level standard errors describe split sensitivity and do not establish generalization to new families. The partition-level thresholds were fixed before the runs and are interpreted at that level.

\section{Reproducibility details}
\label{app:reproduce}

\subsection{Evidence reported in the paper}

The primary quantitative result is a paired comparison of ranking-loss-on and ranking-loss-off controllers across 330 family-held-out partitions. Both variants use the same partition seeds and the same scoring convention, so the shared best-fixed reference cancels in their paired difference. The 330 partitions reuse 82 task families, and the reported standard error measures sensitivity to partition composition, not uncertainty over new families.

An earlier controller-versus-baseline comparison was withdrawn. Its scorer used a 25\% validation fraction while the controllers were trained with 20\%, producing reporting-family sets that did not match. A preregistered split-aligned reanalysis replaced it. The corrected comparison (+0.0092 over the per-split best fixed action, 120 partitions, three guards checked before any value was read) is reported in the Results.

\subsection{Primary ablation configuration}

Table~\ref{tab:hyper} gives the centered-target configuration for the separate 120-partition controller-versus-baseline evaluation. The primary 330-partition ranking-loss ablation uses the uncentered-target comparison described in Appendix~\ref{app:experimental_setup}. Counterfactual labels combine Qwen3-8B and Qwen3-14B continuation policies. The utility weights are $\lambda=1$ and $\eta=1$, and task families are partitioned 60/20/20 into train, validation, and test sets. Each legal forced action has one continuation rollout at temperature zero. The merged pool contains 1,036 source-specific checkpoints from 378 task IDs and 82 families, and 138 task IDs occur under both label sources. The two ablation variants differ only in ranking-loss weight (1 versus 0) and both use uncentred targets. Figure~\ref{fig:training} shows the within-run correlation between loss and decision utility over all 120 runs, median -0.082 and negative on 70, with 36 runs improving their decision while their loss rose. Selecting a checkpoint on loss would therefore pick the wrong one on a large minority of runs, which is why the trainer selects on validation decision utility.

The commands and thresholds for every comparison were fixed in a pre-registration before the results were read.

\subsection{Reproduction scope}

Offline scoring can be recomputed from the stored counterfactual outcomes without rerunning any environment branch. Closed-loop results use ToolSandbox's native Overall Score and are a separate experiment, and reproducing them requires the environment and a served policy.

Policies are served with vLLM \citep{kwon2023vllm}, and the labels are produced by the same served policy that executes the episode.

\subsection{Use of AI assistants}

AI assistants were used for implementation, analysis, and manuscript drafting under human direction. The authors are responsible for the claims, data, citations, and final manuscript.

\end{document}